\documentclass[10pt]{article}

\usepackage{microtype}
\usepackage{graphicx}
\usepackage{subfigure}
\usepackage{booktabs} 
\usepackage{color, xcolor}
\usepackage{enumitem}
\usepackage{multirow}
\usepackage{dblfloatfix}    
\usepackage{amsmath}

\usepackage{hyperref}

\usepackage[accepted]{icml2019}

\icmltitlerunning{The Difficulty of Training Sparse Neural Networks}

\begin{document}

\twocolumn[
\icmltitle{The Difficulty of Training Sparse Neural Networks}



\icmlsetsymbol{equal}{*}

\begin{icmlauthorlist}
\icmlauthor{Utku Evci}{res,go}
\icmlauthor{Fabian Pedregosa}{go}
\icmlauthor{Aidan N. Gomez}{go,ox,for}
\icmlauthor{Erich Elsen}{deep}
\end{icmlauthorlist}

\icmlaffiliation{res}{This work was completed as part of the Google AI Residency}
\icmlaffiliation{go}{Google Research, Brain team}
\icmlaffiliation{deep}{Deepmind}
\icmlaffiliation{ox}{University of Oxford}
\icmlaffiliation{for}{for.ai}
\icmlcorrespondingauthor{Utku Evci}{evcu@google.com}

\icmlkeywords{Machine Learning, sparse networks}

\vskip 0.3in
]



\printAffiliationsAndNotice{}  

\begin{abstract}
We investigate the difficulties of training sparse neural networks and make new observations about optimization dynamics and the energy landscape within the sparse regime. Recent work of \citet{Gale2019, Liu2018} has shown that sparse ResNet-50 architectures trained on ImageNet-2012 dataset converge to solutions that are significantly worse than those found by pruning. We show that, despite the failure of optimizers, there is a linear path with a monotonically decreasing objective from the initialization to the ``good'' solution. Additionally, our attempts to find a decreasing objective path from ``bad'' solutions to the ``good'' ones in the sparse subspace fail. However, if we allow the path to traverse the dense subspace, then we consistently find a path between two solutions. These findings suggest that traversing extra dimensions may be needed to escape stationary points found in the sparse subspace.
\end{abstract}

\section{Introduction}
\label{intro}
Reducing parameter footprint and inference latency of machine learning models is an active area of research, fostered by diverse applications like mobile vision and on-device intelligence. Sparse networks, that is, neural networks in which a large subset of the model parameters are zero, have emerged as one of the leading approaches for reducing model parameter count. It has been shown empirically that deep neural networks can achieve state-of-the-art results under high levels of sparsity \citep{han2015learning, louizos2017bayesian, Gale2019} , and this property has been leveraged to significantly reduce the parameter footprint and inference complexity \citep{kalchbrenner2018} of densely connected neural networks. However, pruning-based sparse solutions require to train densely connected networks and uses the same, or even greater, computational resources compared to fully dense training, which imposes an upper limit on the size of sparse networks we can train.



\begin{figure}[t]
\centerline{\includegraphics[width=0.9\columnwidth]{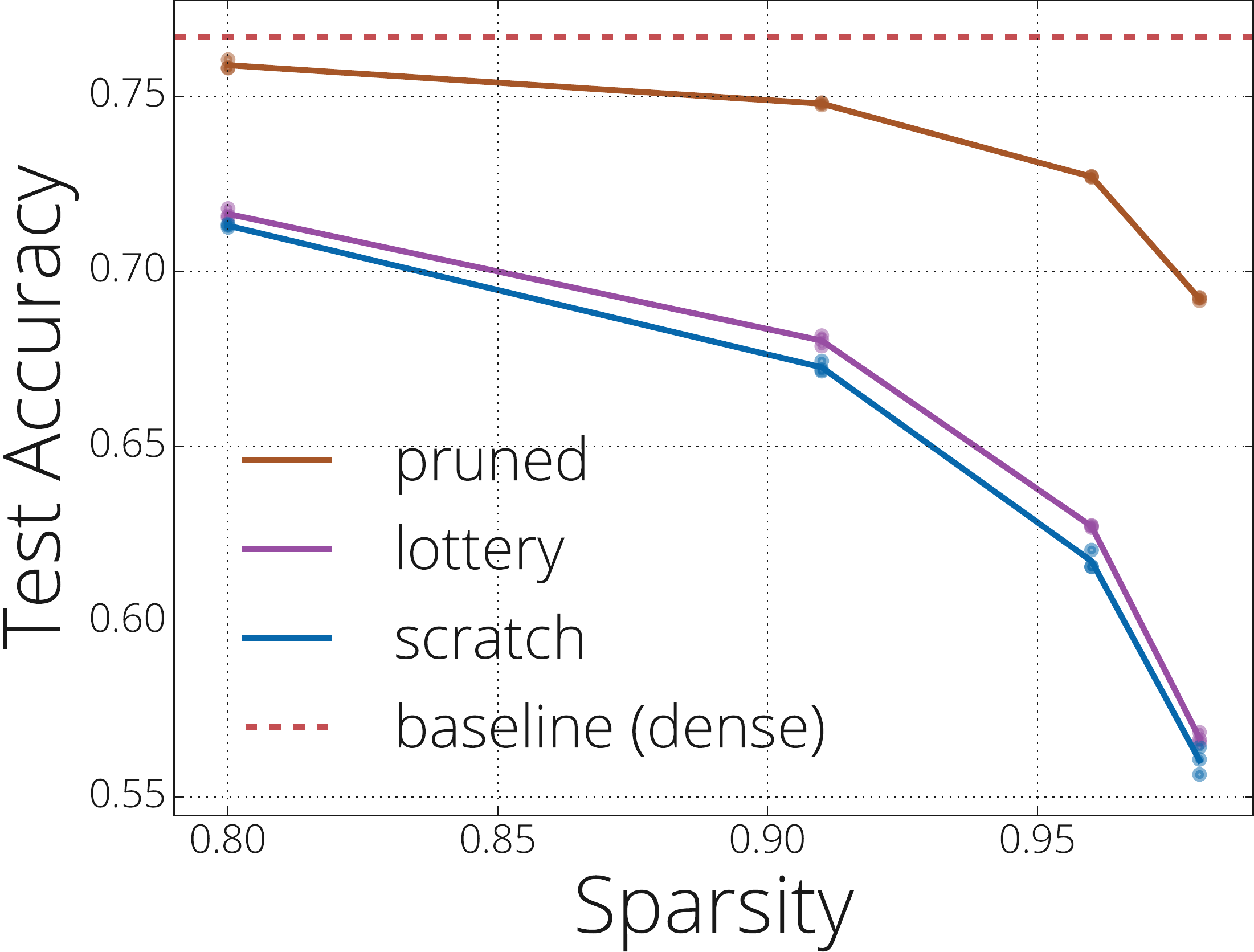}}
\caption{{\bfseries Test accuracy of ResNet-50 networks trained on ImageNet-2012 dataset at different sparsity levels}.
 We observe a large gap in generalization accuracy between approaches based on pruning and other approaches. See text for details. 
 \label{fig:val-acc}}
\vskip -0.2in
\end{figure}

{\bfseries Training Sparse Networks.} In the context of sparse networks, state-of-the-art results have been obtained through training densely connected networks and modifying their topology during training through a technique known as pruning \cite{gupta2018, baiduexploringsparsity, han2015learning}. 
A different approach is to \textit{reuse} the sparsity pattern found through pruning and train a sparse network from scratch. This can be done with a random initialization ({\bfseries ``scratch''}) or the same initialization as the original training ({\bfseries ``lottery''}). Previous work \cite{Gale2019, Liu2018} demonstrated that both approaches achieve similar final accuracies, but lower than pruning\footnote{\citet{frankle2019} closes this gap by using trained parameters from 5th epoch to initialize the network. However they don't quantify how different these values are from the solution.}. The difference between pruning and both approaches to training while sparse can be seen in Figure \ref{fig:val-acc}. 
Despite being in the same energy landscape, ``scratch'' and ``lottery'' solutions fail to match the performance of the solutions found by pruning. Given the utility of being able to train sparse from scratch, it is critical to understand the reasons behind the failure of current techniques at training sparse neural networks.

There exists a line of work on training sparse networks \cite{Bellec2017, Mocanu2018, Pieterse2019, Mostafa2019} which allows the connectivity pattern to change over time.  These techniques generally achieve higher accuracy compared with fixed sparse connectivity, but generally worse than pruning. In this work we focus on \textit{fixed} sparsity patterns, although our results (Section \ref{section32})
also give insight into the success of these other approaches.

Motivated by the disparity in accuracy observed in Figure~\ref{fig:val-acc}, we perform a series of experiments to improve our understanding of the difficulties present in training sparse neural networks, and identify possible directions for future work.

More precisely, our {\bfseries main contributions} are:
\begin{itemize}[leftmargin=*]
\item A set of experiments showing that the objective function is monotonically decreasing along the straight lines that interpolate from:
\begin{itemize}
    \item the original dense initialization\,,
    \item the original dense initialization projected into the sparse subspace \,,
    \item a random initialization in the sparse subspace\,,
\end{itemize}
to the solution obtained by pruning\footnote{sparse subspace refers to the sparsity pattern found by pruning and same in all settings.}. This demonstrates that even when the optimization process fails, there was a monotonically decreasing path to the ``good'' solution. 
\item In contrast, the linear path between the \textit{scratch} and the \textit{pruned} solutions depicts a high energy barrier between the two solutions. Our attempts to find quadratic and cubic B\'ezier curves \cite{Garipov2018} with a decreasing objective between the two sparse solutions fails suggesting that the optimization process gets attracted into a ``bad'' local minima.
\item Finally, by removing the sparsity constraint from the path, we are consistently able to find decreasing objective B\'ezier curves between the two sparse solutions. This result suggests that allowing for dense connectivity might be necessary and sufficient to escape the stationary point converged in the sparse subspace.  
\end{itemize}

The rest of the paper is organized as follows: In \S\ref{methods}, we describe
the experimental setup. In \S\ref{results} we present the results from these experiments, followed by a discussion in \S\ref{scs:conclusion}.

\section{Methods}
\label{methods}
\begin{figure}[t]
\begin{center}
\centerline{\includegraphics[width=0.9\columnwidth]{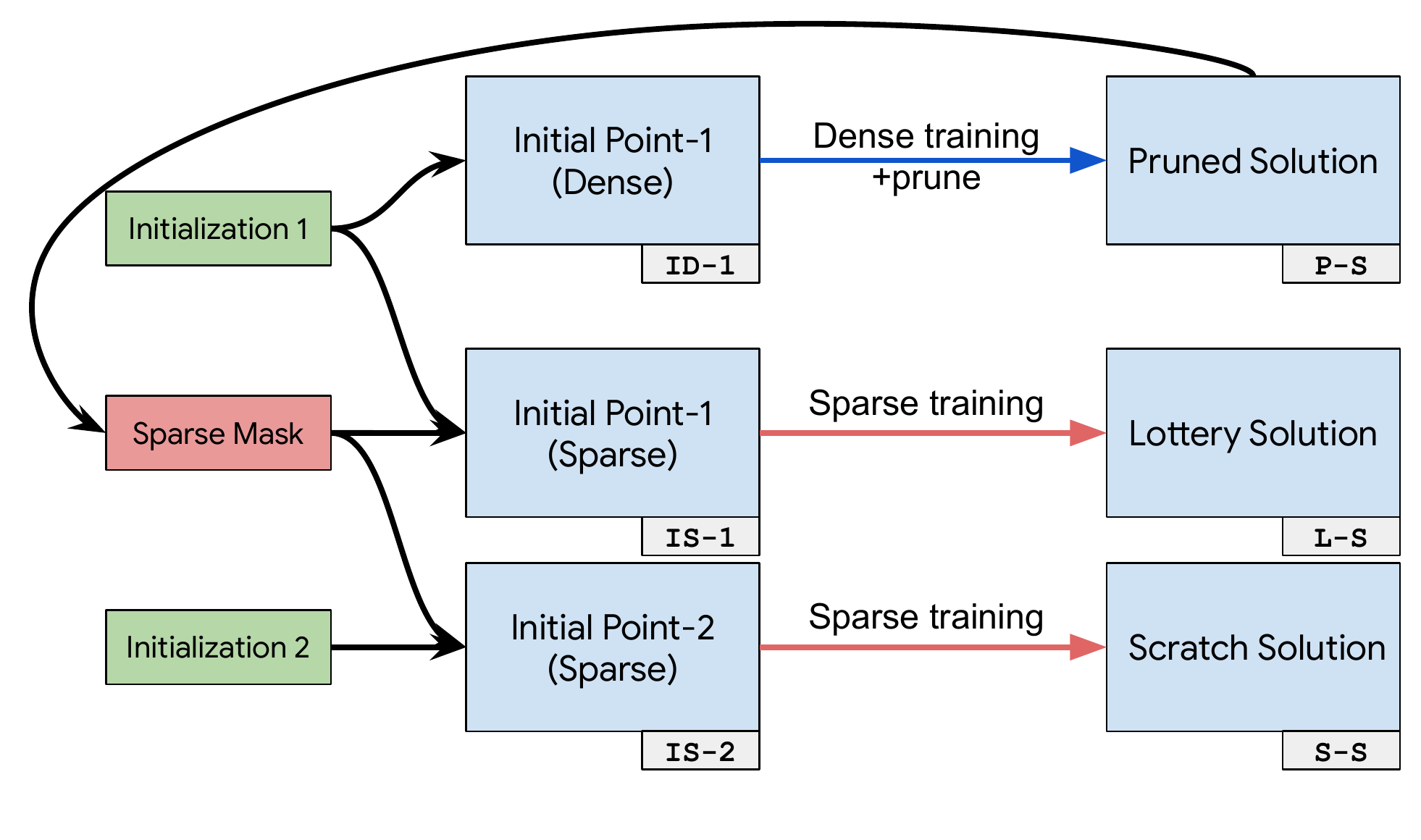}}
\caption{{\bfseries Experimental setup}. In this paper we consider three different methods for obtaining sparse solutions: pruned, lottery and scratch. The pruned solution is obtained by starting with a densely connected network and gradually removing connections during training, whereas the other two solutions are obtained by training sparse networks from start. \label{fig:diagram}}
\end{center}
\vskip -0.3in
\end{figure}

\textbf{Training methods.} The different training strategies considered in this paper are summarized in Figure \ref{fig:diagram}.
In \textit{dense training}, we train the densely connected model and apply model pruning \citep{gupta2018} to find the \textbf{Pruned Solution (P-S)}.
The other strategies start instead with a sparse connectivity pattern (represented as a binary mask) obtained from the pruned solution.
The solution obtained from the same random initialization as the pruned solution \citep{frankle2018} is denoted \textbf{Lottery Solution (L-S)}, while the solution obtained from another random initialization \citep{Liu2018} is named \textbf{Scratch Solution (S-S)}. All of our experiments are based on the Resnet-50 architecture \cite{resnet} and the Imagenet-2012 dataset \cite{imagenet}. Abbreviations defined in Figure \ref{fig:diagram} below the boxes are re-used to in the remaining of the text to indicate start and end points of the interpolation experiments.


\textbf{Pruning strategy.} We use magnitude based model pruning~\cite{gupta2018} in our experiments. This has been shown \cite{Gale2019} to perform as well as the more complex and computationally demanding variational dropout \cite{Molchanov2017} and $\ell_0$ regularization approaches \cite{Louizos2018}. In our experiments, we choose the 3 top performing pruning schedules for each sparsity level using the code and checkpoints provided by \citet{Gale2019}. 
The hyper-parameters involved in the pruning algorithm were found by grid search separately for each sparsity level. The 80\% sparse model loses almost no accuracy over the baseline, while the 98\% sparse model drops to 69\% top-1 accuracy (see Figure \ref{fig:val-acc}-\textit{pruned}). Training details and the exact pruning schedules used in our experiments are detailed in Appendix \ref{apx:expdetail}.

\begin{figure*}[ht]
\begin{center}
\centerline{\includegraphics[width=1.8\columnwidth]{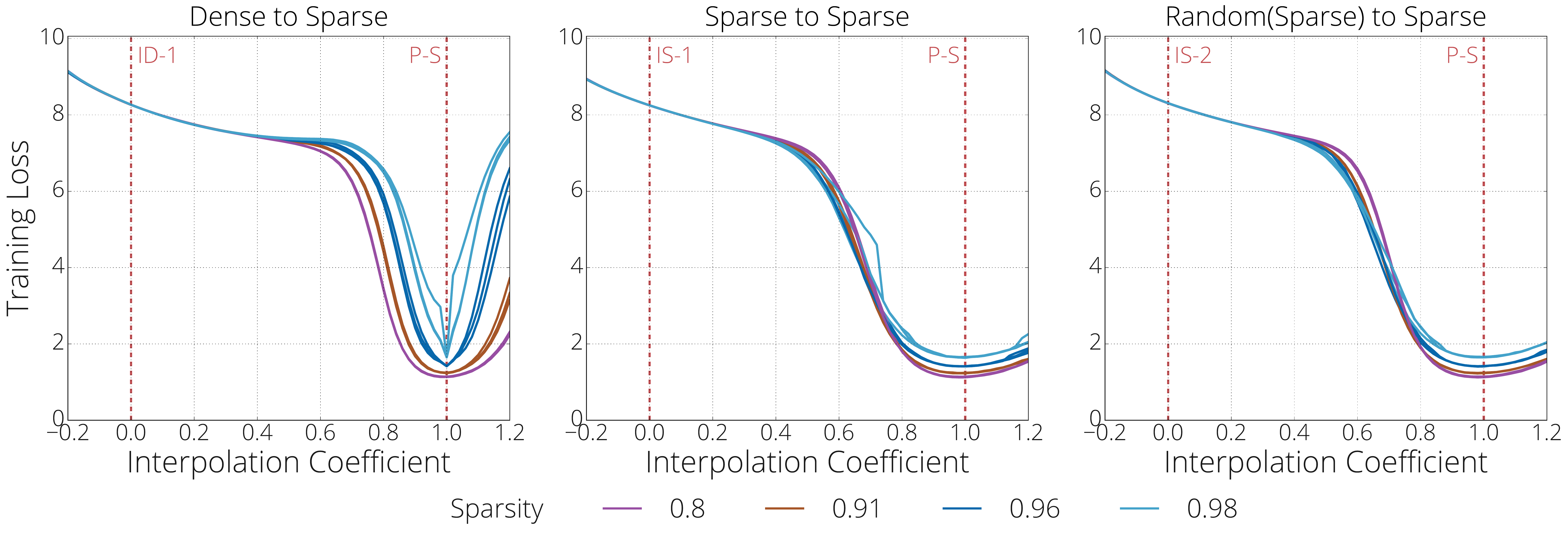}}
\caption{{\bfseries Linear interpolation experiments} between various initial and final points. Interpolations are created with 0.02 increments and evaluated on 500k data-augmented images from training set. Initial and final points (corresponding to coefficients 0 and 1 respectively) are labeled with abbreviations as presented in Figure \ref{fig:diagram}. From all points considered there exist a monotonically decreasing path to the solution found through pruning. \label{fig:interpolation-linear}}
\end{center}
\vskip -0.4in
\end{figure*}
\textbf{Interpolation in parameter space.} Visualizing the energy landscape of neural network training is an active area of research. \citet{Goodfellow2014} measured the training loss on MNIST~\cite{mnist} along the line segment between the initial point $\mathbf{\theta_s}$ and the solution $\mathbf{\theta_e}$, observing a monotonically decreasing curve. Motivated by this, we were curious if this was still true (a) for Resnet-50 on Imagenet-2012 dataset and (b) if it was still true in the sparse subspace. We hypothesized that if (a) was true but (b) was not, then this could help explain some of the training difficulties encountered with sparse networks.

In our linear interpolation experiments, we generate networks along the segment $\mathbf{\theta}=t\mathbf{\theta_e}+(1-t)\mathbf{\theta_s}$ for $t \in [-0.2, 1.2]$ with increments of $0.01$ and evaluate them on the training set of 500k images. Interpolated parameters include the weights, biases and trainable batch normalization parameters. We enable training mode for batch normalization layers so that the batch statistics are used during the evaluation. The objective is identical to the objective used during training, which includes a weight decay term scaled by $10^{-4}$.

We now seek to find a non-linear path between the initial point and solution using parametric B\'ezier curves of order $n = 2$ and $3$. These are curves given by the expression
\vskip -0.2in
\[
 B_n(t)=\sum_{i=0}^n \binom{n}{i} (1-t)^{n-i}t^i\theta_i~,
\]
where $\theta_0=\theta_e$ and $\theta_n=\theta_s~$. We optimize the following,
\vskip -0.2in
\[
\min_{\theta_1,\cdots,\theta_{n-1}} \int_0^1 L(B_n(t))dt~,
\]
using the stochastic method as proposed by \citet{Garipov2018} with a batch size of 2048 where $L(\theta)$ denotes the training loss as a function of trainable parameters. Mirroring our original training settings, we set the weight decay coefficient to $10^{-4}$. We performed a hyper-parameter search over base learning rates ($1,10^{-1},10^{-2},10^{-3}$) and momentum coefficients ($0.9, 0.95, 0.99, 0.995$), obtaining similar learning curves for most of the combinations. We choose $0.01$ as the base learning rate and $0.95$ as the momentum coefficient for these path finding experiments.

\section{Results and Discussion}
\label{results}

\begin{figure*}[ht]
\begin{center}
\centerline{\includegraphics[width=1.85\columnwidth]{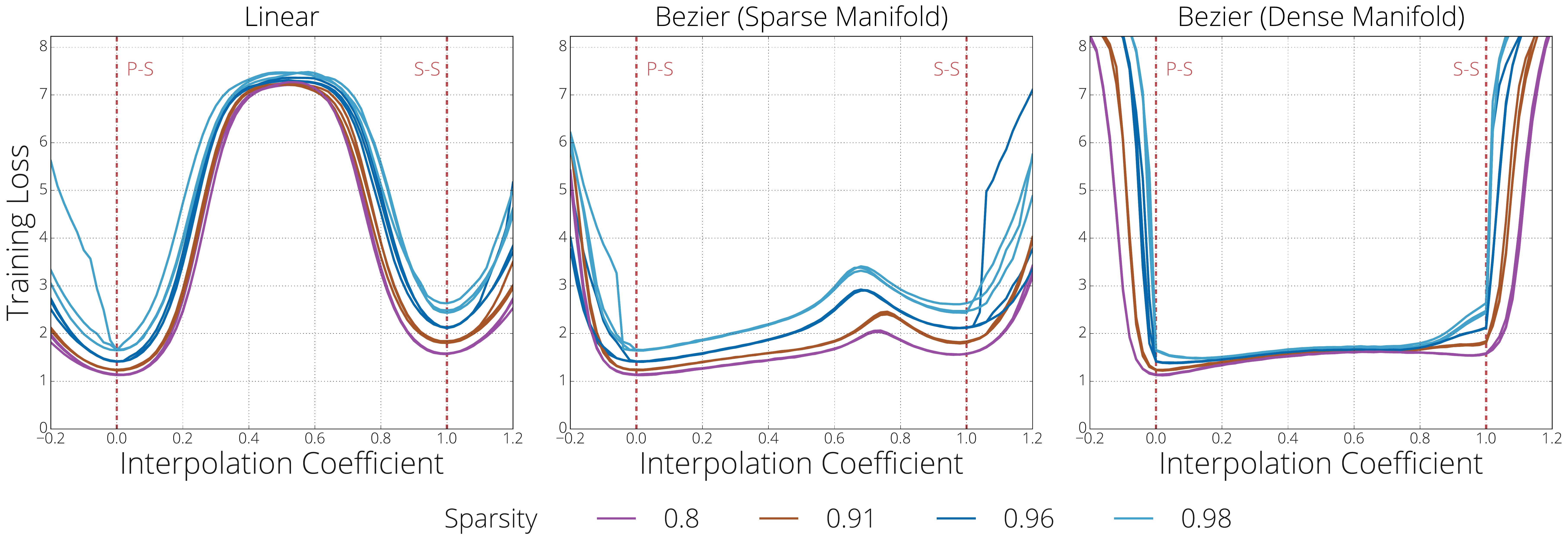}}
\caption{Interpolation experiments between pruned (P-S) and scratch(S-S) sparse solutions: \textbf{(left)} linear interpolation \textbf{(middle, right)} B\'ezier curves minimized in (sparse, dense) manifolds. Loss values are calculated using 500k images from the training set.
\label{fig:interpolation-solution}}
\end{center}
\vskip -0.3in
\end{figure*}

Our experiments highlight a gap in our understanding of energy landscape of sparse deep networks. Why does training a sparse network from scratch gets stuck at a neighborhood of a stationary point
with a significantly higher objective? This is in contrast with recent work that has proven that such a gap does not exist for certain kinds of over-parameterized dense networks \citep{exploringhighdimensionallandscapes2014, lossurfaces2014}. Since during pruning dimensions are slowly removed, we conjecture that this prevents the optimizer from getting stuck into ``bad'' local minima. The failure of the optimizer is even more surprising in the light of the linear interpolation experiments of Section \ref{section31}, which show that sparse initial points are connected to the pruning solutions through a path in which the training loss is monotonically decreasing. 


In high dimensional energy landscapes, it is difficult to assess whether the training converges to a local minimum or to a higher order saddle point. \citet{sagun2017} shows that the Hessian of a convolutional network trained on MNIST is degenerate and most of its eigenvalues are very close to zero indicating an extremely flat landscape at solution. \citet{dauphin2014} comments on \citet{Bray2007}'s results and argues that critical points that are far from the global minima in Gaussian fields are most likely to be saddle points. In Section \ref{section32}, we examine the linear interpolation between solutions and attempt to find a parametric curve between them with decreasing loss. This is because finding a decreasing path from the high loss solution (``scratch'') to the low loss solution(``pruned'') would demonstrate that the former solution is at a saddle point.

\subsection{Path between start and end}
\label{section31}
Linear interpolations from Initial-Point-1 (Dense), Initial-Point-1 (Sparse) and Initial-Point-2 (Sparse) to Pruned Solution at different sparsity levels are shown in Figure \ref{fig:interpolation-linear} respectively; all cases show monotonically decreasing curves. The training loss represented in the y-axis consists of a cross entropy loss and an $\ell_2$ regularization term. While in Figure \ref{fig:interpolation-linear} the $y$ axis represents the full training loss, the two terms composing this loss are shown separately in Appendix \ref{apx:loss-decomp}. There we observe that the sum is dominated by the cross entropy loss.

In Figure \ref{fig:interpolation-linear}-left, we observe a long flat plateau followed by a sharp drop: this is unlike typical learning curves, which are steepest at the beginning and then level off. Model pruning allows the optimizer to take the path of steepest descent while still allowing it to find a good solution as dimensions are slowly removed.


Finally, the linear interpolation from a random point sampled from the original initialization distribution (``scratch'') also depicts a decreasing curve (Figure \ref{fig:interpolation-linear}-right), almost identical to the interpolations that originates from the lottery initialization (Figure \ref{fig:interpolation-linear}-middle). This brings further evidence against the ``lottery ticket'' initialization being special relative to other initializations.
\subsection{Path between two solutions}
\label{section32}
The training loss along the linear segment and the parametric B\'ezier curve connecting the scratch and the pruned solutions are shown in Figure \ref{fig:interpolation-solution}. As observed by \citet{Keskar2016}, linear interpolation (Figure \ref{fig:interpolation-solution}-left) depicts a barrier between solutions, as high as the values observed by randomly initialized networks. The sparse parametric curve (Figure \ref{fig:interpolation-solution}-middle) found through optimization also fails at connecting the two solutions with a monotonically decreasing path (although it has much smaller loss value than the straight line).
Using a third order B\'ezier curve also fails to decrease the maximum loss value over the second order curve (Appendix \ref{apx:loss-bezier3}).  The failure of the third order curve does not prove that a path cannot be found. However, as a second order curve was sufficient to connect solutions in dense networks \cite{Garipov2018}, it does show that if such a path exists, then it must be significantly more complex than those necessary in dense networks.


We continue our experiments by removing the sparsity constraint from the quadratic B\'ezier curve and optimize over the full parameter space (Figure \ref{fig:interpolation-solution}-right). With all dimensions unmasked, our algorithm consistently finds paths along which the objective is significantly smaller\footnote{While this path is not strictly monotonically decreasing, this is not unexpected, given that our algorithm minimizes the integral of the objective over the interpolation segment and so monotonicity is not enforced.}. This result suggests enabling extra connections might be necessary and sufficient to escape   
bad critical points converged. It also gives insight why allowing the sparsity pattern to change over training to be beneficial for training sparse networks.




\section{Conclusion and Future Work}\label{scs:conclusion}
In this work we have provided some insights into the dynamics of optimization in the sparse regime which we hope will guide progress towards better regularization techniques, initialization schema, and/or optimization algorithms for training sparse networks.



Training of sparse neural networks is still not fully understood from an {optimization} perspective.  In the sparse regime, we show that optimizers converge to stationary points with a sub-optimal generalization accuracy. This is despite monotonically decreasing paths existing from the initial point to the pruned solution. And despite nearly monotonically decreasing paths in the dense subspace from the ``bad'' local minimum to the pruned one.

Optimizers that avoid “bad” local minima in the sparse regime are yet to be found. We believe that understanding why popular optimizers used in deep learning fail in the sparse regime will yield important insights leading us towards more robust optimizers.

\subsubsection*{Acknowledgments}
The authors would like to thank; Trevor Gale, Hugo Ponte, Saurabh Kumar, Vincent Dumoulin, Danny Tarlow, Ross Goroshin, Pablo Castro, Nicolas Le Roux, Levent Sagun and Kevin Swersky for helpful discussions and feedback.

\bibliographystyle{icml2019}
\bibliography{index}

\clearpage
\appendix
\onecolumn

\section{Experimental details}
\label{apx:expdetail}
Our experiments use the code made publicly available by \footnote{\url{https://github.com/google-research/google-research/tree/master/state_of_sparsity}}. The pruning algorithm uses magnitude based iterative strategy to reach to a predefined final sparsity goal over the coarse of the training. We use a batch size of 4096 and train the network with 48000 steps (around 153 epochs). Our learning rate starts from 0 and increases linearly towards 0.1 in 5 epoch and stays there until 50th epoch. The learning rate is dropped by a factor of 10 afterwards at 50th, 120th and 140th epochs \cite{Goyal2017}. 

\begin{table}[h]
\caption{Pruning strategies used in various settings. }
\label{table:1}
\vskip 0.15in
\begin{center}
\begin{small}
\begin{sc}
\begin{tabular}{c|c|c|c}
Sparsity & Start & End & Frequency \\
\hline
\multirow{3}{*}{0.8} & 12500 & 40000 & 2000 \\
& 12500 & 40000 & 500 \\
& 12500 & 36000 & 1000 \\
\hline
\multirow{3}{*}{0.91} & 12500 & 36000 & 2000 \\
& 10000 & 40000 & 4000 \\
& 7500 & 36000 & 1000 \\
\hline
\multirow{3}{*}{0.96} & 7500 & 36000 & 1000 \\
& 10000 & 32000 & 500 \\
& 7500 & 40000 & 2000 \\
\hline
\multirow{3}{*}{0.98} & 12500 & 36000 & 500 \\
& 7500 & 32000 & 500 \\
& 12500 & 32000 & 500 \\
\end{tabular}
\end{sc}
\end{small}
\end{center}
\vskip -0.1in
\end{table}

Due to high sensitivity observed, we don't prune the first convolutional layer and cap the maximum sparsity for the final fully connected layer with 80\%. Top 3 performing pruning schedules selected for each sparsity level are shared in Table \ref{table:1}. We calculate the average $\ell_1$ norm of the gradient for the first setting in the table and obtain 4e-6. As is the case for most iterative (especially stochastic) methods, our solutions do not qualify as stationary points since the gradient is never exactly zero. By slight abuse of language we refer to stationary point as any point where the gradient is below of $10^{-5}$. 

\begin{figure}[h!]
\begin{center}
\centerline{\includegraphics[width=.5\linewidth]{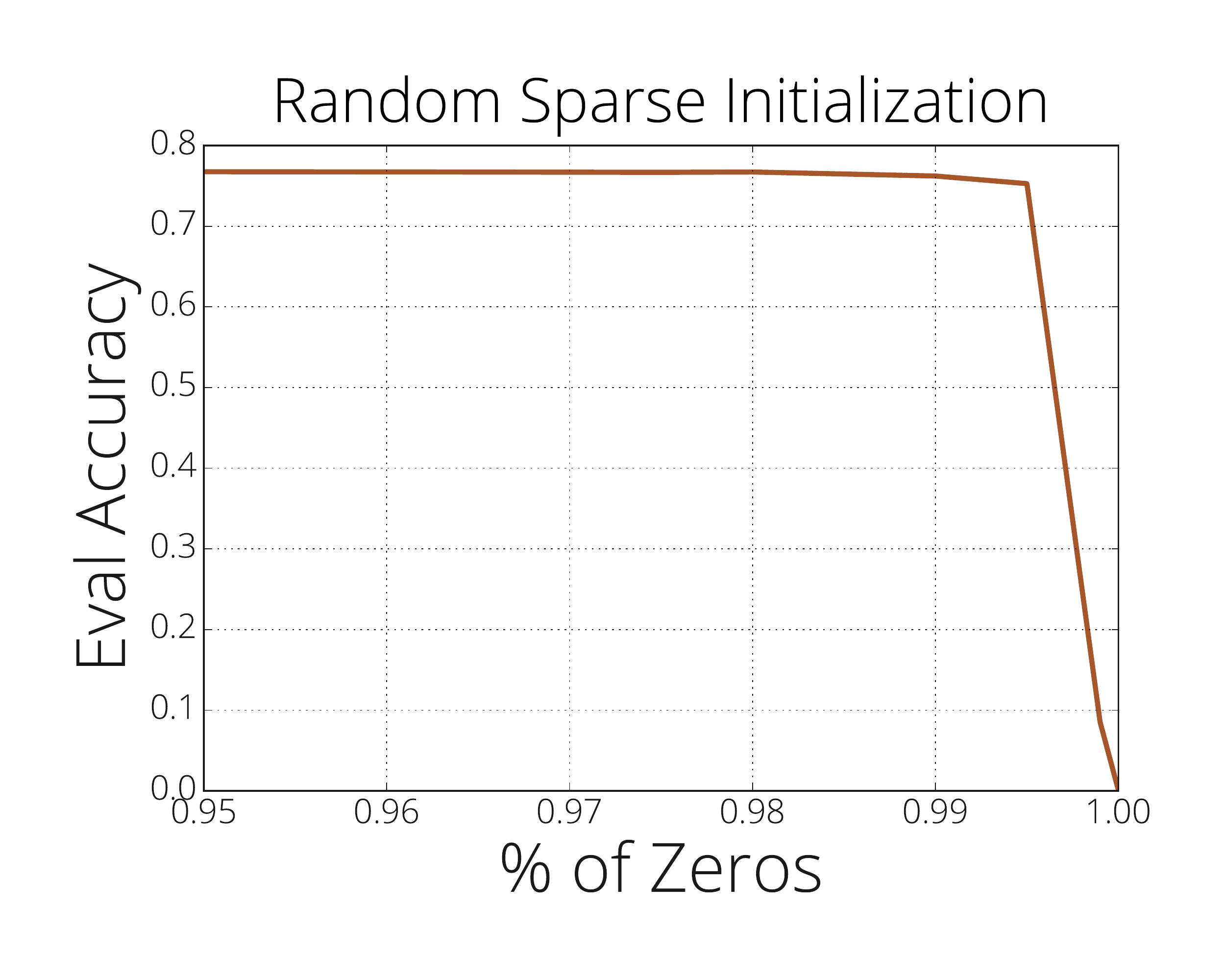}}
\vskip -0.2in
\caption{At the beginning of the training we randomly set a fraction of weights to zero and train the network with default parameters for 32k steps. We observe a sudden drop only if more than 99\% of the parameters are set to zero. \label{fig:sparse-init}}\end{center}
\vskip -0.3in
\end{figure}

\section{Sparse Initialization}
\label{apx:sparse-init}
Initialization methods that control variance of activations during the first forward and backward-pass is known to be crucial for training deep neural networks \cite{glorot2010, he2015}. However, with batch normalization \cite{Ioffe2015} and skip connections the importance of initialization is expected to be less pronounced. \textit{sparse-init} experiments shared in Figure \ref{fig:val-acc} can be seen as a demonstration of such tolerance. In sparse-init experiments we train a dense ResNet-50 but use the sparse binary mask found by pruning to set a fraction of initial weights to zero. At all sparsity levels considered (0.8, 0.91, 0.96, 0.98), we observe that the training succeeds and reaches to final accuracy around 76\% matching the performance of the original training. Thus the initialization point alone cannot be the reason for the failure of sparse training.

To understand the extent which we are able to train sparsely initialized networks without compromising performance, we perform experiments where we randomly set a given fraction of weights to zero. Figure \ref{fig:sparse-init} shows the results. We observe no significant difference until 99.5\% after which we observe a sharp drop in performance. The initialization requires a very small number of non-zero weights to succeed.


\section{Loss Decomposition}
\label{apx:loss-decomp}
\begin{figure*}[h]
\centerline{\includegraphics[width=\linewidth]{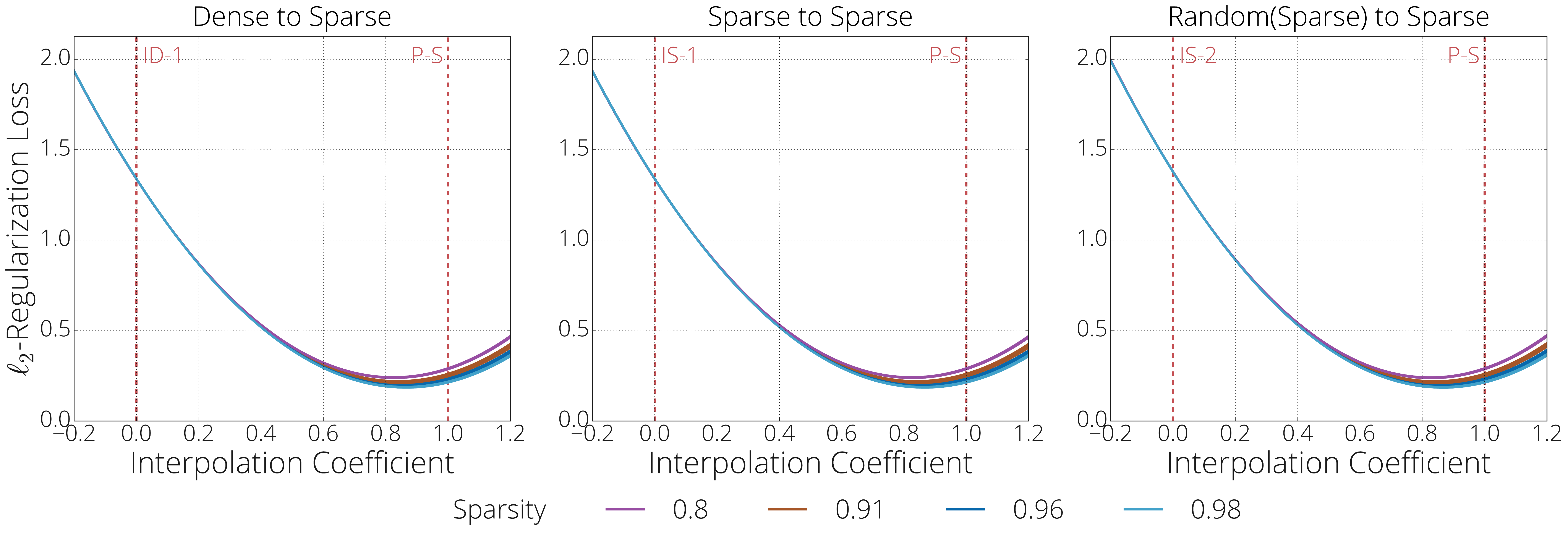}}
\caption{Regularization term of the linear interpolations of Figure \ref{fig:interpolation-linear}.\label{fig:interpolation-reg_loss}}
\end{figure*}

Figure~\ref{fig:interpolation-reg_loss} depicts the value of the \(\ell_2\) regularization term over the linear interpolations described in Section~\ref{section31}. The curve demonstrates that the solutions found are consistently of lower weight magnitude than their initialization, and they also demonstrate that the regularization terms are a factor of ten smaller than the objective function.

\begin{figure*}[h]
\begin{center}
\centerline{\includegraphics[width=\linewidth]{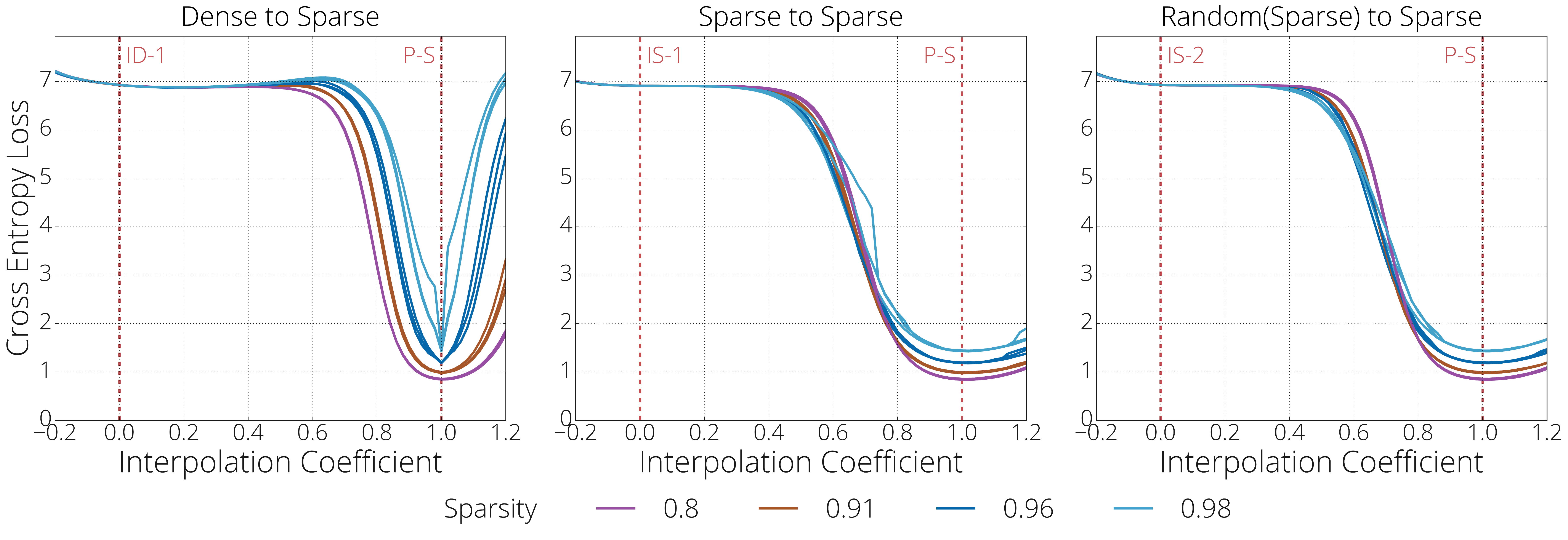}}
\caption{Cross entropy term of the linear interpolations of Figure \ref{fig:interpolation-linear}. \label{fig:interpolation-cross_loss}}
\end{center}
\vskip -0.3in
\end{figure*}

Figure~\ref{fig:interpolation-cross_loss} depicts the value of the cross entropy loss over the linear interpolation described above. Figure~\ref{fig:interpolation-cross_loss}-(left) the sparse to sparse and random sparse to sparse interpolation maintain the monotonic decreasing pattern observed in the interpolations plots of the total loss (Figure~\ref{fig:interpolation-linear}), whereas dense to sparse interpolation shows a slight increase in the objective before a rapid descent.

Dense-sparse cross entropy is increasing. Time to drop is much less with original initializations, higher with random and least with dense.

\section{Path Finding Experiments with Cubic B\'ezier Curve}
\label{apx:loss-bezier3}
\begin{figure*}[h]
\begin{center}
\centerline{\includegraphics[width=\linewidth]{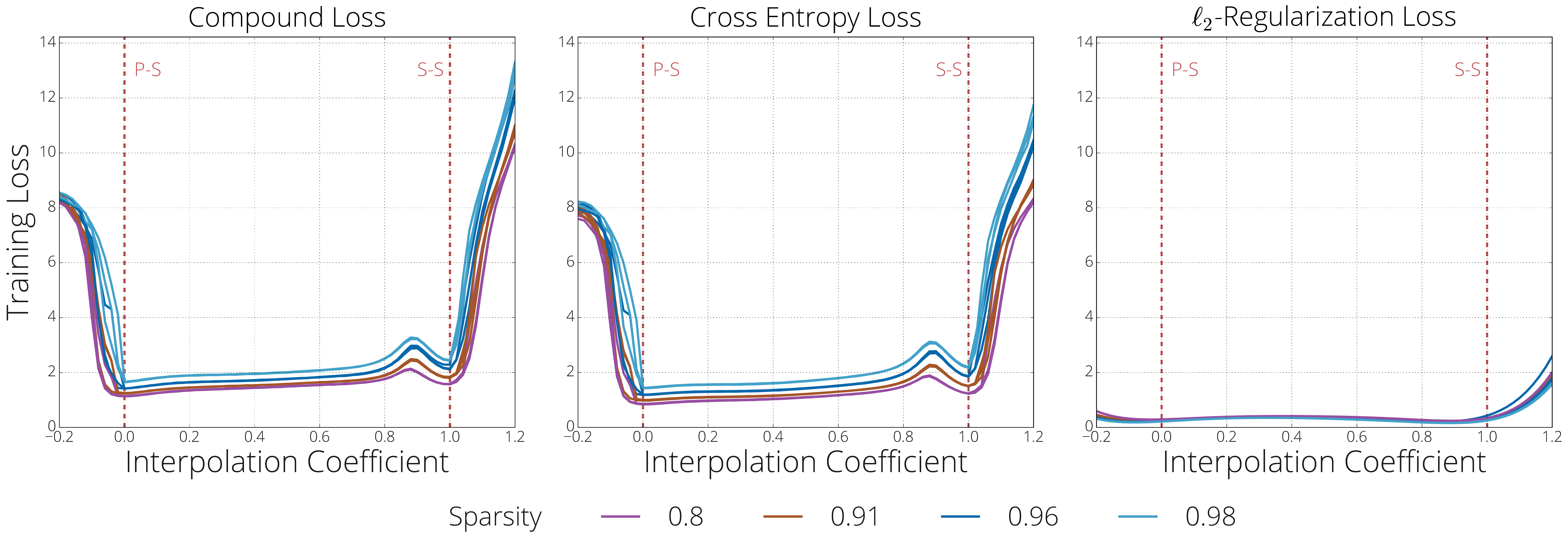}}
\caption{Training loss along the third order (cubic) B\'ezier curve found between the ``pruning'' and ``scratch'' solutions.\label{fig:bezier3_loss}}
\end{center}
\vskip -0.3in
\end{figure*}

In Section \ref{section32} our experiments fail to find paths along which the loss is decreasing between the ``pruned'' and ``scratch'' solutions. Would optimizing more complex parametric curves change the result? Figure~\ref{fig:bezier3_loss} depicts the objective along the third order B\'ezier curves. Though the integral of the loss over the segment between solutions seems less than Figure \ref{fig:interpolation-solution}-middle, we still observe a small barrier between solutions. 

\end{document}